\documentclass[letterpaper, 10 pt, journal, twoside]{IEEEtran}
\usepackage{url}
\usepackage{cite}
\usepackage{amsmath,amssymb,amsfonts}
\usepackage{algorithmic}
\usepackage{graphicx}
\usepackage{textcomp}
\usepackage{xcolor}
\usepackage{algorithm}
\usepackage{caption}
\usepackage{float} 
\usepackage{graphbox}
\usepackage{tcolorbox}
\usepackage[export]{adjustbox}
\usepackage{subcaption}
\usepackage{bibentry}
\usepackage{bm}
\usepackage{multirow} 
\usepackage{makecell}
\usepackage{threeparttable}

\ifCLASSINFOpdf
\else
\fi

\begin{document}
%
\title{Seamless Virtual Reality\\with Integrated Synchronizer and Synthesizer\\for Autonomous Driving}
%
%
%

\author{He Li$^{1}$, Ruihua Han$^{3,4}$, Zirui Zhao$^{3}$, Wei Xu$^{5}$, Qi Hao$^{3}$, Shuai Wang$^{2,\dagger}$, and Chengzhong Xu$^{1,\dagger}$%
\thanks{Manuscript received: November 8, 2023; Revised January 27, 2024; Accepted February 26, 2024. This paper was recommended for publication by Editor Ashis Banerjee upon evaluation of the Associate Editor and Reviewers' comments.
This work was supported by the Science and Technology Development Fund of Macao S.A.R (FDCT) (No. 0123/2022/AFJ and No. 0081/2022/A2), the National Natural Science Foundation of China (No. 62371444 and No. 6226116065), and Shenzhen Fundamental Research Program (No. JCYJ20220818103006012).}

\thanks{$^{1}$He Li and Chengzhong Xu are with the State Key Laboratory of Internet of Things for Smart City (SKL-IOTSC), University of Macau, Macau, China {\tt\footnotesize \{mc25094, czxu\}@um.edu.mo}}%
\thanks{$^{2} $Shuai Wang is with the Center for Cloud Computing, Shenzhen Institute of Advanced Technology (SIAT), Chinese Academy of Sciences, Shenzhen, China {\tt\footnotesize s.wang@siat.ac.cn}}%
\thanks{$^{3, 4} $Ruihua Han is with the Department of Computer Science and Engineering, Southern University of Science and Technology, Shenzhen, Guangdong, China, and also with the Department of Computer Science, The University of Hong Kong, Hong Kong {\tt\footnotesize hanrh@connect.hku.hk}}%
\thanks{$^{3} $Zirui Zhao and Qi Hao are with the Department of Computer Science and Engineering, Southern University of Science and Technology, Shenzhen, China {\tt\footnotesize zhaozr@mail.sustech.edu.cn, hao.q@sustech.edu.cn}}%
\thanks{$^{5} $Wei Xu is with the Manifold Tech Limited, Hong Kong, China {\tt\footnotesize xuwei@manifoldtech.cn}}%
\thanks{$^\dagger$ denotes the corresponding authors}
\thanks{Digital Object Identifier (DOI): see top of this page.}
}
%
%

\markboth{IEEE Robotics and Automation Letters. Preprint Version. Accepted February, 2024}
{Li \MakeLowercase{\textit{et al.}}: Seamless Virtual Reality with Integrated
Synchronizer and Synthesizer for Autonomous Driving} 

%



\makeatletter
\g@addto@macro\@maketitle{
\centering
\setcounter{figure}{0}
\vspace{0.1in}
\begin{center}
    \centering
    \includegraphics[width=1\textwidth]{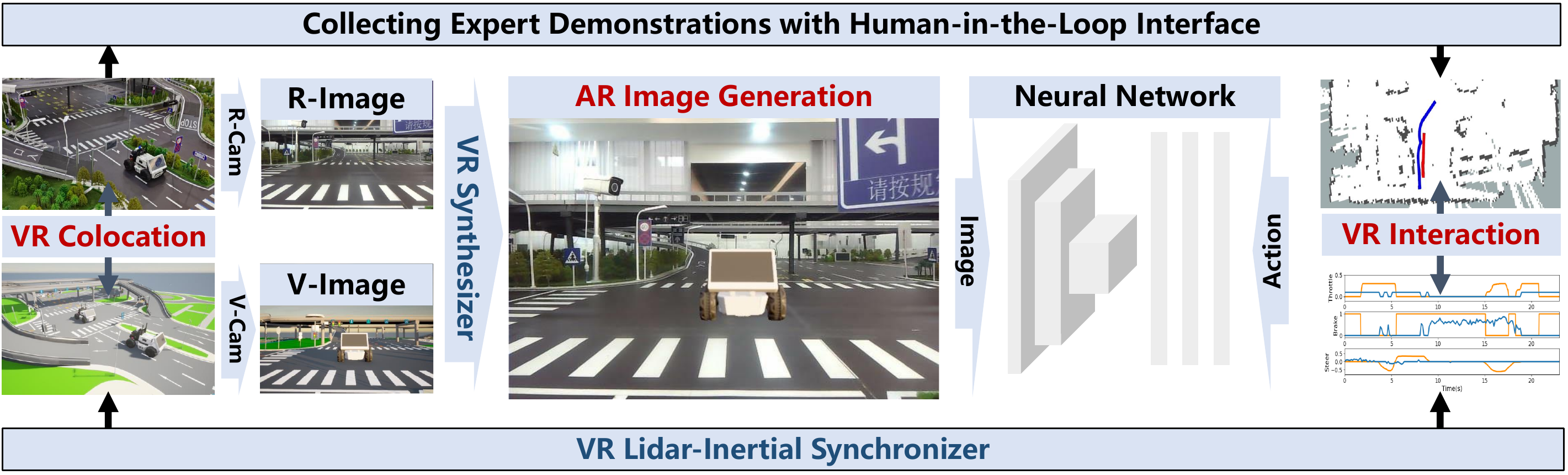}
\captionof{figure}{
Illustration of $\mathsf{SVR}$ with $\mathsf{IS}^2$. A vehicle navigates in the physical space while interacting with obstacle vehicles in the digital space. The $\mathsf{SVR}$ generated dataset is used to train AD neural networks.
}\label{fig:eyecatcher}
\end{center}
\vspace{-0.2in}
}
\makeatother
\maketitle

\begin{abstract}
Virtual reality (VR) is a promising data engine for autonomous driving (AD). However, data fidelity in this paradigm is often degraded by VR inconsistency, for which the existing VR approaches become ineffective, as they ignore the inter-dependency between low-level VR synchronizer designs (i.e., data collector) and high-level VR synthesizer designs (i.e., data processor). This paper presents a seamless virtual reality ($\mathsf{SVR}$) platform for AD, which mitigates such inconsistency, enabling VR agents to interact with each other in a shared symbiotic world. 
The crux to $\mathsf{SVR}$ is an integrated synchronizer and synthesizer ($\mathsf{IS}^2$) design, which consists of a drift-aware lidar-inertial synchronizer for VR colocation and a motion-aware deep visual synthesis network for augmented reality image generation.
We implement $\mathsf{SVR}$ on car-like robots in two sandbox platforms, achieving a cm-level VR colocalization accuracy and $3.2\%$ VR image deviation, thereby avoiding missed collisions or model clippings.
Experiments show that the proposed $\mathsf{SVR}$ reduces the intervention times, missed turns, and failure rates compared to other benchmarks. The $\mathsf{SVR}$-trained neural network can handle unseen situations in real-world environments, by leveraging its knowledge learnt from the VR space.
\end{abstract}


%
\IEEEpeerreviewmaketitle

\section{Introduction}
%
%
%
%
\IEEEPARstart{S}{carcity} of boundary data (e.g., cut-in, crash) in real-world environments has led to incompleteness of the datasets used for design and validation of autonomous driving (AD) systems \cite{scarcity,tesla,nvidia}. 
This makes it a challenge to push AD from closed to open scenarios.
Compared to road tests, virtual simulation \cite{carla,flcav,tesla,nvidia,V2XSIM}, which is based on modern computer graphics and physical modeling technologies, is a safer, faster, and cheaper method that can generate boundary scenarios that are difficult to be produced in physical environments. However, there exists a non-negligible gap between the simulated and real-world data \cite{scarcity,aads,flcav}. 

To mitigate the gap, a promising solution is to leverage virtual reality (VR) (i.e., defined as a pair of physical and digital worlds in this paper) for more realistic data generation. 
Conventional VRAD schemes \cite{translation,surfelgan,aads} process off-the-shelf datasets using some sim-to-real transfer methods, which fail in collecting end-to-end data (e.g., no action-in-the-loop).
Emerging interactive VRAD \cite{dense-rl,safety-ar,opencda-ros,carla_bridge} leverages vehicle-in-the-loop (VIL) or environment-in-the-loop (EIL) to collect end-to-end augmented reality (AR) data, while ensuring exact vehicle dynamics and road environments.
Nonetheless, VIL or EIL approaches may break down 
whenever VR inconsistency emerges, due to low-frequency colocation (e.g. RTK-GPS colocation \cite{dense-rl,safety-ar}), incomplete interaction (e.g., no EIL \cite{opencda-ros,carla_bridge}), or lack of reciprocal considerations between low-level VR designs (i.e., data collector) and high-level VR designs (i.e., data processor) \cite{translation,surfelgan,dense-rl,safety-ar,opencda-ros,carla_bridge}. Such inconsistency would result in degraded data fidelity, e.g., missed collisions or model clippings.

\begin{figure*}[t]
    \centering
    \includegraphics[width=0.98\textwidth]{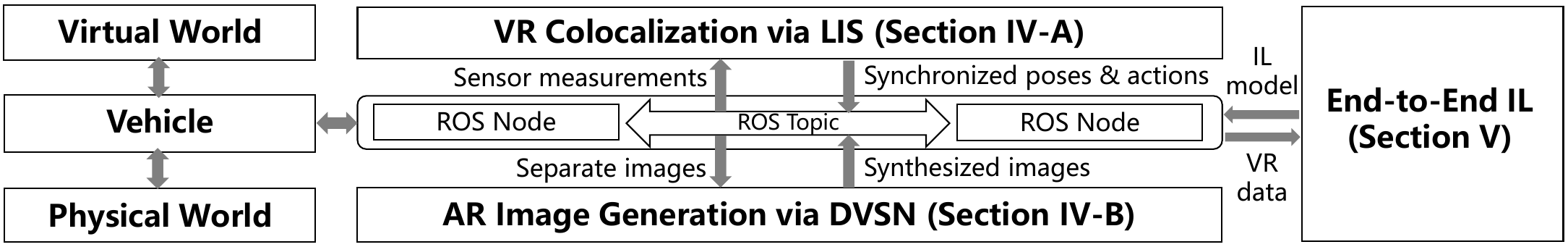}
    \caption{System architecture of $\mathsf{SVR}$, which integrates $\mathsf{LIS}$, $\mathsf{DVSN}$, and $\mathsf{IL}$, through ROS.}
    \label{fig:system}
    \vspace{-0.2in}
\end{figure*}

To fill in the blank, this paper presents a seamless virtual reality ($\mathsf{SVR}$) platform with integrated synchronizer and synthesizer ($\mathsf{IS}^2$) to maximize the VR data fidelity (as shown in Fig.~1). 
We first propose a drift-aware lidar-inertial synchronizer ($\mathsf{LIS}$) based on regularized error-state Kalman filter for VR colocation. With $\mathsf{LIS}$, a pair of aligned VR poses (i.e., left hand side of Fig.~1) and the associated control vectors (i.e., right hand side of Fig.~1) are simultaneously collected at high frequency.
To obtain aligned VR video streams and augmented images, this paper further proposes a motion-aware deep visual synthesis network 
($\mathsf{DVSN}$), which leverages neighbourhood consensus network (NC-Net) for image registration and segment anything model (SAM) for image aggregation. 
The $\mathsf{DVSN}$ effectively separates front objects from background in the virtual images and automatically calibrates objects' positions and sizes so as to best merge them into the real images for AR image generation (i.e., middle of Fig.~1).
Various experiments in robot operation system (ROS) show that $\mathsf{LIS}$ and $\mathsf{DVSN}$ achieve cm-level colocalization errors and minimum image deviations, enabling vehicles to simultaneously interact with physical and digital objects smoothly in a symbiotic $\mathsf{SVR}$ world. Experiments on a passenger vehicle in a parking lot are also provided to validate the effectiveness of $\mathsf{SVR}$ in real-world settings.

Furthermore, while VR data can augment the AD dataset, limited experimental evaluations have been reported for VR-trained imitation learning (IL) (a widely used AD paradigm). To this end, a human-in-the-loop interface (i.e., upper of Fig.~1) is used to collect expert driving datasets, and a conditional IL network is trained and tested (i.e., middle of Fig.~1) on car-like robots in two sandbox platforms. 
Experiments show that the proposed $\mathsf{SVR}$ trained IL reduces the intervention times, missed turns, and failure rates compared to benchmarks trained by real data, virtual simulation, or conventional VR methods. 
Interestingly, it is found that the $\mathsf{SVR}$-trained IL can handle unseen situations in real environments, by leveraging its knowledge learnt from the VR space.

Our contributions are summarized as follows: 
\begin{itemize}
    \item Propose a drift-aware $\mathsf{LIS}$, which accurately maps autonomous vehicles in the real world and adversarial obstacles in the virtual world to a symbiotic world; 
    \item Propose a motion-aware $\mathsf{DVSN}$, which automatically calibrates and segments objects from virtual images for seamless synthesis with environments from real images;
    \item Implementation of $\mathsf{SVR}$-trained IL and evaluation of the performance gain brought by $\mathsf{SVR}$ to IL.
\end{itemize}

%

\section{Related Work}

\textbf{Virtual simulation} can generate corner cases that are difficult to be collected in the real world \cite{carla}. This technique adopts unreal engine (UE) for photo-realistic rendering (e.g., ray-casting) and physical engine (PE) for dynamics modeling (e.g., crash).
Various AD simulators have been released, e.g., Intel Carla, ROS Gazebo, NVIDIA Drive, Waymo CarCraft, Tesla Autopilot, LG SVL, Tecent TAD-SIM, THU DAIR-V2X, UCLA OpenCDA, SJTU V2X-SIM, SIAT CarlaFLCAV \cite{carla,flcav,V2XSIM}. 
Nevertheless, the gap between real and virtual worlds is non-negligible, which significantly impedes the credibility of AD simulation.

\textbf{VRAD}, which integrates VR and AD techniques, have been adopted to mitigate the above gap.
VRAD can be categorized into non-interactive \cite{translation,surfelgan,aads} and interactive approaches \cite{carla_bridge,opencda-ros,safety-ar,dense-rl}. 
Non-interactive VRAD approaches \cite{translation,surfelgan,aads} focus on processing off-the-shelf datasets instead of generating interactive datasets, which are suitable for perception data augmentation instead of end-to-end data augmentation.
For instance, generative adversarial networks (GAN) \cite{translation} have been adopted for domain adaption. 
Waymo and Google developed SurfelGAN, which can generate VR driving videos from real historical clips \cite{surfelgan}. 
Baidu proposed augmented autonomous driving simulation (AADS) system, which generates augmented sensor data under different illuminations and weathers from AplloScape \cite{aads}.

\textbf{Interactive VRAD}, on the other hand, leads to a paradigm shift where previous isolated virtual and real agents can now naturally cooperate and compete in a shared driving scenario \cite{xr}.
While interactive VR has been studied in various human-robot interaction tasks \cite{xr,iGibson}, researches on interactive VRAD are still in its infancy. 
Recently, Intel, UCLA, and UMICH have released interactive VRAD platforms.
For instance, Intel released Carla-ROS-Bridge interfaces for VIL simulation \cite{carla_bridge}, 
and UCLA further bridged Carla-ROS with OpenCDA \cite{opencda-ros}.
UMICH developed a VRAD platform that allows real vehicles in the physical test track to interact with virtual obstacles \cite{safety-ar}, and adopted this platform to train a reinforcement learning network for AD \cite{dense-rl}. 

Nonetheless, the above interactive VRAD systems involve VR inconsistency as mentioned in Section I.
Here, we present an $\mathsf{IS}^2$ approach that achieves $\mathsf{SVR}$ by joint design of low-level VR (i.e., data collector) and high-level VR (i.e., data processor). 
We also demonstrate the effectiveness of $\mathsf{SVR}$ for IL with extensive real-world experiments, and for the first time, the performance gain brought by VRAD to IL is concisely quantified.

\section{The $\mathsf{SVR}$ System}
The architecture of $\mathsf{SVR}$ is shown in Fig.~\ref{fig:system}, which adopts the $\mathsf{LIS}$ for VR colocation, the $\mathsf{DVSN}$ for AR image generation, and the ROS interface for collecting end-to-end interactive datasets to train the IL network.

The $\mathsf{LIS}$ module simultaneously estimates the vehicle poses ${^V\mathbf{s}}_t=({^{V}x}_{t},{^{V}y}_{t},{^{V}\theta}_{t})$ and ${^R\mathbf{s}}_t=({^{R}x}_{t},{^{R}y}_{t},{^{R}\theta}_{t})$ in the virtual and real spaces, respectively, where $t$ is the index of lidar time frame, $(x_{t},y_{t})$ and $\theta_{t}$ are positions and orientations, of the virtual vehicle (if the upper-left subscript is $V$) and the real vehicle (if the upper-left subscript is $R$). The two poses are connected through a rigid transformation with the extrinsic parameters of rotation matrix ${^{VR}\mathbf{R}}_t$ and translation vector ${^{VR}\mathbf{t}}_t$ between the two worlds. 
The initial guesses of extrinsics are ${^{VR}\mathbf{R}}_t^{[0]},{^{VR}\mathbf{t}}_t^{[0]}$, which can be obtained using offline calibration. 
The VR colocation needs to update state vectors 
${^V\mathbf{s}}_t,{^R\mathbf{s}}_t$ and their extrinsics ${^{VR}\mathbf{R}}_t,{^{VR}\mathbf{t}}_t$ as the vehicle moves in the VR space.
In our system, this is realized based on lidar measurements ${^V\mathbf{z}}_t,{^R\mathbf{z}}_t$, and inertial measurements ${^V\mathbf{w}}_t,{^R\mathbf{w}}_t$ at the virtual and real vehicles.
As such, the colocation algorithm can operate at $50$ Hz even at low-cost vehicles in GPS-denied environments.

With colocation results from $\mathsf{LIS}$, a sequence of paired images 
${^V\mathcal{U}}=\{{^V\mathbf{u}}_1,{^V\mathbf{u}}_2,\cdots\}$ and 
${^R\mathcal{U}}=\{{^R\mathbf{u}}_1,{^R\mathbf{u}}_2,\cdots\}$, as well as corresponding real-world actions 
$\mathcal{A}=\{\mathbf{a}_1,\mathbf{a}_2,\cdots\}$, are obtained, where $\mathbf{u}_k\in\mathbb{R}^{LW\times 1}$ denotes the real image (if the upper-left subscript is $R$) and its associated virtual image (if the upper-left subscript is $V$) of the $k$-th image frame with $L$ and $W$ being the image length and width, and $\mathbf{a}_k$ denotes the action vector containing steer, throttle, and brake elements.
The $\mathsf{DVSN}$ module defines a mapping that fuses virtual image ${^V\mathbf{u}}_k$ and real image ${^R\mathbf{u}}_k$ into augmented image ${^A\mathbf{u}}_k=f_{\mathsf{DVSN}}({^V\mathbf{u}}_k,{^R\mathbf{u}}_k)$.
The resultant end-to-end interactive dataset is given by 
${^A\mathcal{U}}=\{{^A\mathbf{u}}_1,{^A\mathbf{u}}_2,\cdots\}$ and $\mathcal{A}=\{\mathbf{a}_1,\mathbf{a}_2,\cdots\}$.

The IL module is an inference mapping 
$f_{\mathsf{IL}}: \{{^R\mathcal{U}},\mathcal{C}\} \to \mathcal{A}$
that maps real images and branch indicators into actions, where $\mathcal{C}=\{{c}_1,{c}_2,\cdots\}$, with $c_k$ being a high-level command and acting as a switch that selects a certain back-end network
At the $k$-th image frame, we have $\mathbf{a}_k=f_{\mathsf{IL}}({^R\mathbf{u}}_k,c_k|\mathbf{y}^*)$, where $\mathbf{y}^*$ is the IL weights trained with the expert demonstrations.
In conventional AD systems, 
the expert demonstrations are collected from the real-world environment and $\mathbf{y}$ is trained on the real-world dataset $\{{^R\mathcal{U}},{\mathcal{A}}\}$.
In our system, however, 
the expert demonstrations are collected in the VR space and
$\mathbf{y}^*$ is obtained from the VR dataset $\{{^A\mathcal{U}},{\mathcal{A}}\}$.

\section{The $\mathsf{IS}^2$ Approach}

\subsection{VR Colocation with Drift-Aware $\mathsf{LIS}$}

The $\mathsf{LIS}$ uses IMU measurements and features extracted in two consecutive virtual-reality scans to estimate the relative transformation of the vehicle and the extrinsic parameters between the VR space. Let ${^R\mathbf{x}}_t$ denote the vehicle state transformation from lidar time-step $t$ to $t+1$ in the real world:
\begin{align}
&{^R\mathbf{x}}_{t,t+1}
=\left[
{^R\mathbf{p}}_{t,t+1},
{^R\mathbf{v}}_{t,t+1}
{^R\mathbf{q}}_{t,t+1},  
{^R\mathbf{b}}_a,
{^R\mathbf{b}}_g,
{^R\mathbf{g}}_t
\right],
\end{align}
where ${^R\mathbf{p}}_{t,t+1}$ and ${^R\mathbf{q}}_{t,t+1}$ represent the translation and quaternion from time $t$ to $t+1$, 
${^R\mathbf{v}}_{t,t+1}$ is the velocity,
$^R\mathbf{b}_a$ is the acceleration bias,
$^R\mathbf{b}_g$ is the gyroscope bias, and $^R\mathbf{g}_t$ is the local gravity. 
All the above states are defined w.r.t. the local frame (i.e., IMU-affixed frame). 
We adopt error-state representation and the error vector of ${^R\mathbf{x}}_{t,t+1}$ is given by
\begin{align}
&\delta\mathbf{x}
=\left[
{\delta\mathbf{p}},
{\delta\mathbf{v}},
{\delta\theta},
{\delta\mathbf{b}}_a,
{\delta\mathbf{b}}_g,
{\delta\mathbf{g}}
\right], \label{state}
\end{align}
where ${\delta\theta}$ is the error of angle ${^R\theta}_{t,t+1}$ associated with ${^R\mathbf{q}}_{t,t+1}$. 
With \eqref{state}, 
we compute the state posterior ${^R\mathbf{x}}_{t,t+1}^{\mathrm{post}}$ by injecting the error term $\delta\mathbf{x}$ into the state prior ${^R\mathbf{x}}_{t,t+1}^{\mathrm{pri}}$ as:
\begin{align}
{^R\mathbf{x}}_{t,t+1}^{\mathrm{post}}
&={^R\mathbf{x}}_{t,t+1}^{\mathrm{pri}}
\boxplus
\delta\mathbf{x}
\nonumber\\
&
=\Big[
{^R\mathbf{p}}_{t,t+1}
^{\mathrm{pri}}
+{\delta\mathbf{p}}
,\,
{^R\mathbf{v}}_{t,t+1}
^{\mathrm{pri}}+
{\delta\mathbf{v}},
\,
{^R\mathbf{q}}_{t,t+1}^{\mathrm{pri}}
\otimes
\mathrm{exp}(\delta\theta), 
\nonumber\\
&\quad\quad{}{}
{^R\mathbf{b}}_a^{\mathrm{pri}}
+{\delta\mathbf{b}}_a,
{^R\mathbf{b}}_g^{\mathrm{pri}}
+{\delta\mathbf{b}}_g,
{^R\mathbf{g}}_t^{\mathrm{pri}}
+{\delta\mathbf{g}}
\Big], \label{xpost}
\end{align}
where $\boxplus$ is defined in the second line of equation \eqref{xpost}, $\otimes$ is the quaternion product, and $\mathrm{exp}$ maps the angle to quaternion.

\begin{figure}[t]
    \centering
    \includegraphics[height=0.17\textwidth]{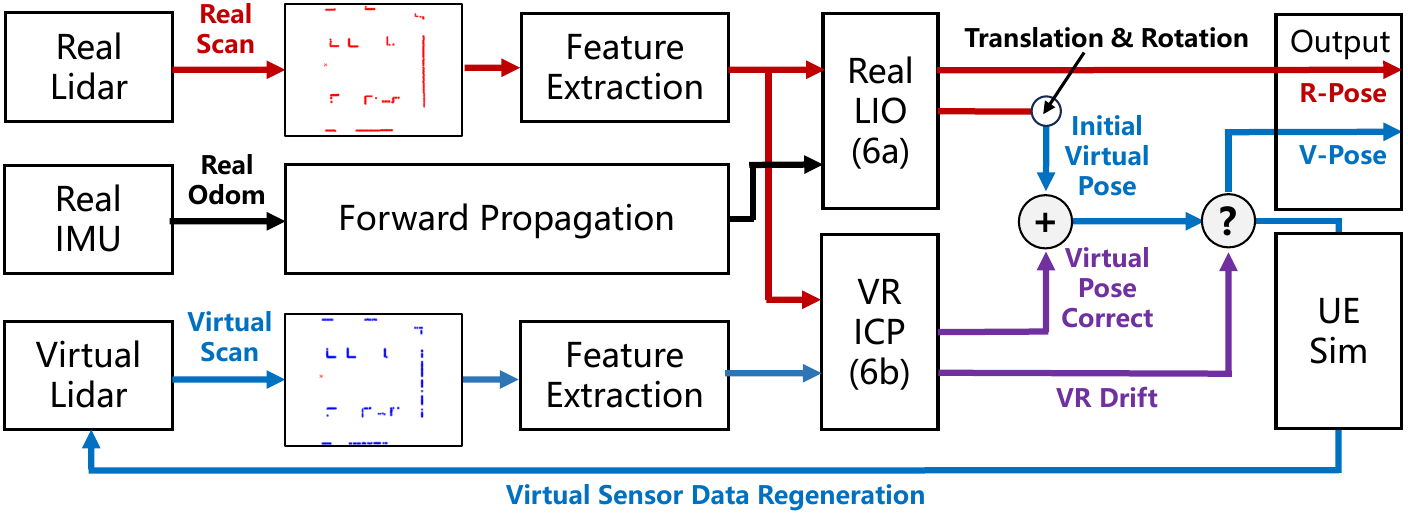}
    \caption{Structure of $\mathsf{LIS}$ for VR colocation}
    \vspace{-0.05in}
    \label{LIS}
\end{figure}

Based on the above analysis, the key of $\mathsf{LIS}$ is to estimate the error term $\delta\mathbf{x}$. 
However, in contrast to existing state estimation methods \cite{lio,ijrr} that only match the error term with the real-world sensor measurements, 
the VR colocation needs to further incorporate the virtual measurements into the state estimation for VR drift computations. 
This is because the goal of VR colocation is to collect synchronized VR measurements instead of accurate localization. 
To this end, we present a VR iterated Kalman filter approach shown in Fig.~\ref{LIS}, 
where its forward propagation that outputs state prior is a standard approach (detailed in \cite{lio,ijrr}), but its backward propagation that outputs state posterior is different from existing approaches.

Specifically, since the virtual vehicle is a digital twin of the real vehicle, we have 
${^V\mathbf{x}}_{t,t+1}=
H({^R\mathbf{x}}_{t,t+1})
$, where $H$ is determined by ${^{VR}\mathbf{R}_t}$ and ${^{VR}\mathbf{t}_t}$ as follows:

\begin{equation}
\left\{
\begin{aligned}
{^V\mathbf{p}}_{t,t+1}
&=
{^{VR}\mathbf{R}_t}
{^R\mathbf{p}}_{t,t+1}
+{^{VR}\mathbf{t}_t}, \nonumber\\
{^V\mathbf{v}}_{t,t+1}
&=
{^{VR}\mathbf{R}_t}
{^R\mathbf{v}}_{t,t+1}
+{^{VR}\mathbf{t}_t} , \nonumber\\
{^V\mathbf{q}}_{t,t+1}
&=
{^{VR}\mathbf{q}_t}
\otimes
{^V\mathbf{q}}_{t,t+1},
\end{aligned}
\right.
\end{equation}
where ${^V\mathbf{x}}_{t,t+1}$ is the virtual state transformation, and ${^{VR}\mathbf{q}}_t$ is the quaternion of ${^{VR}\mathbf{R}_t}$.
Then the iterated correction step can be formulated as an optimization problem that minimizes the deviation from the state prior, the residual derived from the real lidar measurement model, and the drift between virtual and real lidar measurements:
    \begin{align}
&\min \limits_{\delta{\mathbf{x}}, {^{VR}\mathbf{R}_t},{^{VR}\mathbf{p}_t}}~~
    \|\delta{\mathbf{x}}\|_{\mathbf{\Lambda}_t}
+\|{^RF}(
{\mathbf{x}}_{t,t+1}^{\mathrm{pri}}
\boxplus
\delta\mathbf{x}
) \|_{({\mathbf{J}}_t{\mathbf{M}}_t{\mathbf{J}}_t^T)^{-1}}
\nonumber\\
&\quad\quad\quad\quad\quad
    +\| {^{VR}F}(
    {\mathbf{x}}_{t,t+1}^{\mathrm{pri}}
\boxplus
\delta\mathbf{x}, 
    H({\mathbf{x}}_{t,t+1}^{\mathrm{pri}}
\boxplus
\delta\mathbf{x}))\|_{2},
\label{problemA}
    \end{align}
where $\|\cdot\|$ is the Mahalanobis norm (weighted Euclidean norm), 
$\mathbf{\Lambda}_t$ is the covariance of $\delta{\mathbf{x}}$, ${\mathbf{J}}_t$ is the Jacobian of $^RF(\cdot)$ w.r.t. the real lidar measurement noise and ${\mathbf{M}}_t$ is the covariance matrix of the real lidar measurement noise.
The noise of virtual lidar measurement is assumed to be zero.
The function ${^RF}(\cdot)$ measures the residual error calculated from real points ${^R\mathbf{z}}_t$ (determined by real state) and real edges computed based on \cite{zhang2014loam}.
The function ${^{VR}F}(\cdot)$ measures the drift error calculated from real points ${^R\mathbf{z}}_t$ and virtual points ${^V\mathbf{z}}_t$  (determined by virtual and real states).

The above problem can be viewed as a drift-aware VR version of the standard Kalman filter problem. The drift term makes the variable $\delta\mathbf{x}$ coupled in multiple norm terms. 
We propose a penalty alternating minimization (PAM) based on variable splitting.
In particular, introduce a copy
$\delta \mathbf{y}=\delta \mathbf{x}$ and 
replace 
$\delta\mathbf{x}$
with $\delta \mathbf{y}$ in the function $^{VR}F(\cdot)$.
The newly introduced equality constraint is transformed into a penalty term
$\rho\|\delta \mathbf{x}-\delta \mathbf{y}\|_{\mathbf{\Lambda}_t}$ ($\|\cdot\|_{\mathbf{\Lambda}_t}$ is adopted to simplify subsequent problem solving) and added to the objective function of \eqref{problemA}, where $\rho$ is a hyper-parameter.
The penalty problem is then solved by an alternating minimization approach that optimizes $\delta\mathbf{x}$ with $\{{^{VR}\mathbf{R}_t},{^{VR}\mathbf{p}_t},
\delta\mathbf{y}\}$ fixed, and 
optimizes $\{{^{VR}\mathbf{R}_t},{^{VR}\mathbf{p}_t},
\delta\mathbf{y}\}$ with $\delta\mathbf{x}$ fixed:
    \begin{subequations}
    \begin{align}
\min \limits_{\delta{\mathbf{x}}}~~&
    \|\delta{\mathbf{x}}\|_{\mathbf{\Lambda}_t}
    +
\|{^RF}(
{\mathbf{x}}_{t,t+1}^{\mathrm{pri}}
\boxplus
\delta\mathbf{x}
) \|_{({\mathbf{J}}_t{\mathbf{M}}_t{\mathbf{J}}_t^T)^{-1}}
\nonumber\\
&
    +\rho\|\delta{\mathbf{x}}-\delta{\mathbf{y}}\|_{\mathbf{\Lambda}_t},
\label{subproblemA}
\\
\min \limits_{\delta{\mathbf{y}},
{^{VR}\mathbf{R}_t},{^{VR}\mathbf{p}_t}}~~&
\| {^{VR}F}(
    {\mathbf{x}}_{t,t+1}^{\mathrm{pri}}
\boxplus
\delta\mathbf{y}, 
    H({\mathbf{x}}_{t,t+1}^{\mathrm{pri}}
\boxplus
\delta\mathbf{y}))\|_{2}
    \nonumber\\
    &+\rho\|\delta{\mathbf{x}}-\delta{\mathbf{y}}\|_{\mathbf{\Lambda}_t},
\label{subproblemB}
    \end{align}
    \end{subequations}
Given an initial 
$\delta{\mathbf{y}}^{[0]},{^{VR}\mathbf{R}}_t^{[0]},{^{VR}\mathbf{t}}_t^{[0]}$, and by iteratively solving \eqref{subproblemA}--\eqref{subproblemB} according to \cite{ijrr} until convergence (in practice we can early stop the iteration), the solution $\delta{\mathbf{x}}^*, {^{VR}\mathbf{R}^*_t},{^{VR}\mathbf{p}^*_t}$ to \eqref{problemA} is obtained.
Finally, we update the covariance $\mathbf{\Lambda}_t$, posterior real state 
${^R\mathbf{x}}_{t,t+1}^{\mathrm{post}}$, and posterior virtual state 
${^V\mathbf{x}}_{t,t+1}^{\mathrm{post}}
=H({^R\mathbf{x}}_{t,t+1}^{\mathrm{post}})
$. With error states in the local frame, we obtain ${^V\mathbf{s}}_t=({^{V}x}_{t},{^{V}y}_{t},{^{V}\theta}_{t})$ and ${^R\mathbf{s}}_t=({^{R}x}_{t},{^{R}y}_{t},{^{R}\theta}_{t})$ in the world frames of virtual and real spaces.

\subsection{Data Generation with Motion-Aware $\mathsf{DVSN}$}

\begin{figure}
    \centering
    \includegraphics[height=0.17\textwidth]{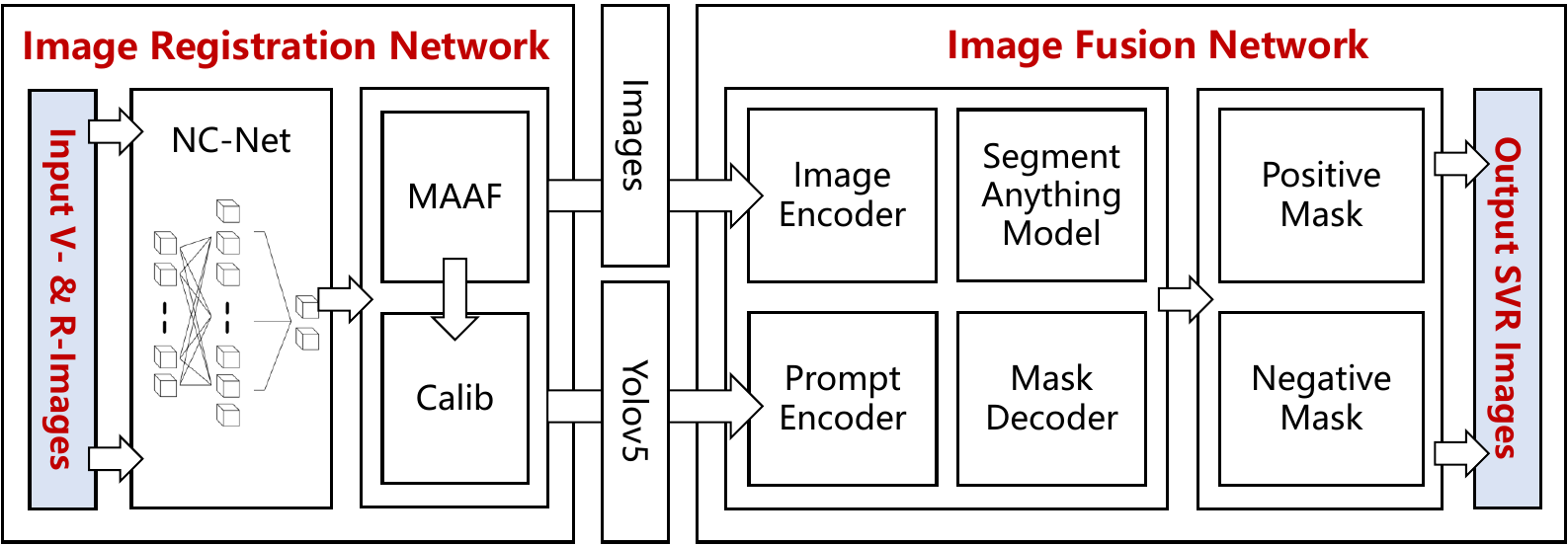}
    \caption{Structure of $\mathsf{DVSN}$ for VR data generation}
    \label{fusion}
\vspace{-0.05in}
\end{figure}

With synchronized vehicle poses in the VR spaces, the VR cameras would generate a pair of images ${^V\mathbf{u}}_k$ and ${^R\mathbf{u}}_k$.
Ideally, we can directly merge ${^V\mathbf{u}}_k$ and ${^R\mathbf{u}}_k$ if their pixels are exactly aligned, 
However, due to practical simulation errors (e.g., inaccurate world, vehicle, camera models), there exist drifts between the virtual and real images, and we need to register them before fusion. Based on this observation, the proposed $\mathsf{DVSN}$ module consists of an image registration network $\mathsf{DVSN}_r(\cdot)$ and an image fusion network $\mathsf{DVSN}_f(\cdot)$, as shown in Fig.~\ref{fusion}.

For $\mathsf{DVSN}_r(\cdot)$, 
it utilizes the neighbourhood consensus networks (NC-Net) \cite{b16} to find a set of matching points between ${^V\mathbf{u}}_k$ and ${^R\mathbf{u}}_k$. 
NC-Net is a convolution neural network that identifies the above correspondences between a pair of images via analyzing neighbourhood consensus patterns in the 4D space. However, there exist incorrect matches due to the uncertainties of NC-Net.\footnote{
Such uncertainties can be mitigated by controlling the input VR drifts below a certain threshold, which is realized by the virtual-to-real point registration in \eqref{subproblemB}.} These mismatched features should be discarded. Here we adopt a motion-aware adaptive filter (MAAF), which discards features by computing a time-varying region of interest (RoI) according to the vehicle motions. 
Specifically, this RoI is defined by a set $\mathbb{G} = \{ \mathbf{g}\in {\mathbb{R}^{2}}| \mathbf{G}\mathbf{g}\leq 
     {\mathbf{d}}\}
$, where 
$\mathbf{G} \in {\mathbb{R}^{l \times 2}}$ and $\mathbf{d} \in {\mathbb{R}^{l}}$, with $l$ being the number of edges for the RoI. 
The center of RoI is denoted as $\mathbf{o}$.
Given the actions $\{\mathbf{a}_{t+h}\}$ and the Ackerman state-action evolution model $E(^R\mathbf{s}_{t+h},\mathbf{a}_{t+h})$ (e.g., defined in equations (8)--(10) of \cite[Sec. III-B]{han2022rda}), the motion-aware RoI needs to minimize the total distance between model predictive trajectories and the ROI center:
    \begin{align}
    &\min \limits_{\mathbf{o}}~\sum^{H-1}_{h=0} 
    \left\|f_{\mathsf{proj}}(^R\mathbf{s}_{t+h})-\mathbf{o}\right\|_2^2 
    \nonumber \\
    &\text {s.t.}~^R\mathbf{s}_{t+h+1}=
E\left(^R\mathbf{s}_{t+h},\mathbf{a}_{t+h}\right),~\forall h, \label{RoI}
\end{align}
where $f_{\mathsf{proj}}$ is the projection from 3D space to 2D image, and $H$ is the length of prediction horizons that can be finetuned.
Denoting the solution to \eqref{RoI} as $\mathbf{o}^*$, the set $\mathbb{G}^*$ is obtained accordingly by varying the interior angle without changing the RoI area, e.g., reshape squares to parallelograms for $l=4$. The MAAF would discard features outside the RoI and the pruned result is used to calculate the perspective transformation $\mathsf{PT}$ on virtual images. 
With $\mathsf{PT}$, we have 
$\mathsf{DVSN}_r({^V\mathbf{u}}_k)=
\mathsf{PT}({^V\mathbf{u}}_k)$.

Next, to merge $\mathsf{DVSN}_r({^V\mathbf{u}}_k)$ and ${^R\mathbf{u}}_k$ for data augmentation, we need to segment the useful objects out of the entire virtual image. 
Here we adopt the segment anything model (SAM) \cite{SAM}, 
which is a state-of-the-art large pre-trained model for image segmentation. 
SAM consists of image encoder, prompt encoder, and mask decoder. 
The image encoder and the prompt encoder can generate image embeddings and prompt embeddings, respectively. 
Then the mask decoder predicts masks using prompt embeddings, image embedding, and an output token with an multi-layer perceptron for foreground probability computation. 
In our $\mathsf{DVSN}$ module, the prompt $\mathcal{P}_k$ for the image $\mathsf{DVSN}_r({^V\mathbf{u}}_k)$ is generated by bounding boxes of YOLOv5 (finetuned with the VR dataset collected from our $\mathsf{SVR}$ platform).
Given prompt $\mathcal{P}_k$, SAM outputs 
negative (for object) and positive (for background) masks ${^N\mathbf{m}}_{k},{^P\mathbf{m}}_{k}\in\{0,1\}^{LW\times 1}$, where values of $1$ in $^N\mathbf{m}_{k}$ indicate pixels that belong to the object and $0$ otherwise.
Denoting $\odot$ as the Hadamard product for element-wise matrix multiplication, the entire procedure of $\mathsf{DVSN}$ can be represented as 
\begin{align}
\mathsf{DVSN}(^V\mathbf{u}_k,{^R\mathbf{u}}_k)&={^N\mathbf{m}}_{k}\odot{^{V}
\mathsf{PT}(\mathbf{u}}_{k})+
{^P\mathbf{m}}_{k}\odot{^R\mathbf{u}}_k.
\end{align}

\section{Experiments}

\subsection{Implementation}

\begin{figure*}[!t]
	\centering
	\begin{subfigure}{0.235\linewidth}
		\centering
		\includegraphics[width=\linewidth]{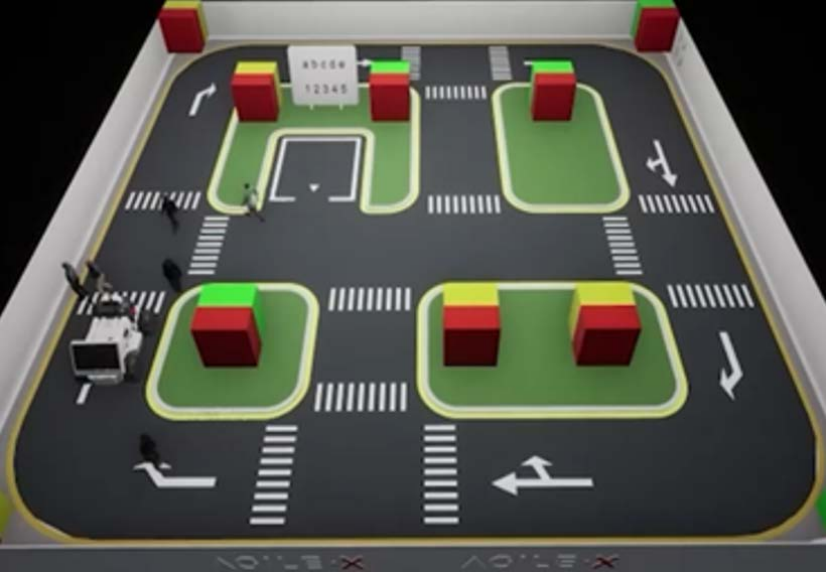}
		\caption{Virtual AGX}
	\end{subfigure}
	\centering
	\begin{subfigure}{0.235\linewidth}
		\centering
		\includegraphics[width=\linewidth]{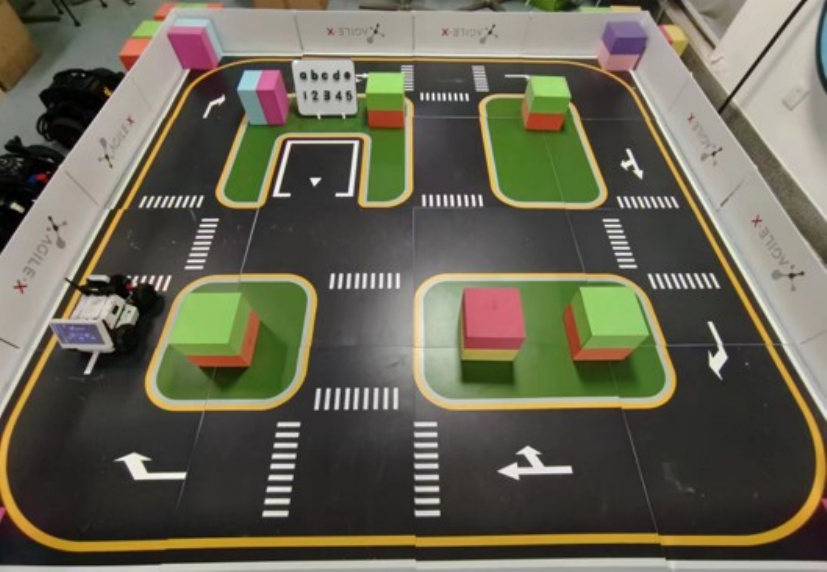}
		\caption{Real AGX}
	\end{subfigure}
 	\begin{subfigure}{0.235\linewidth}
		\centering
		\includegraphics[width=\linewidth]{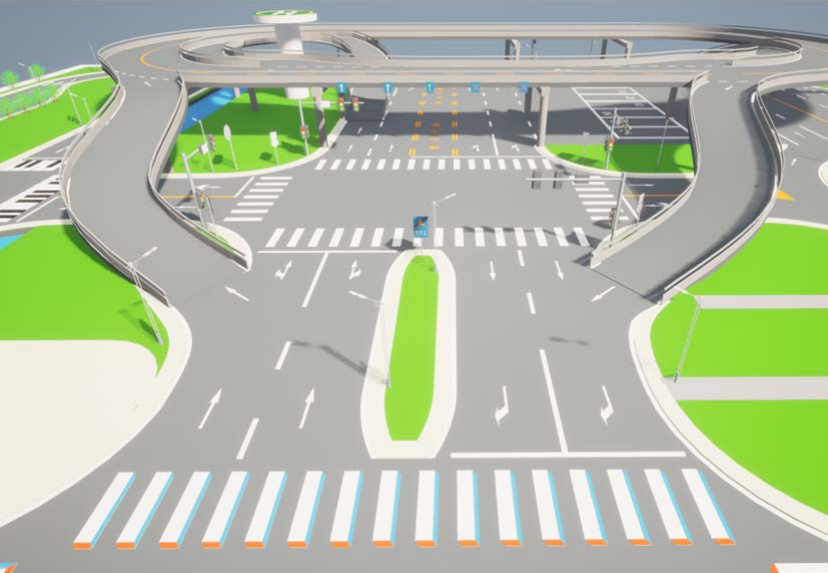}
		\caption{Virtual MoCAM}
	\end{subfigure}
	\centering
	\begin{subfigure}{0.235\linewidth}
		\centering
		\includegraphics[width=\linewidth]{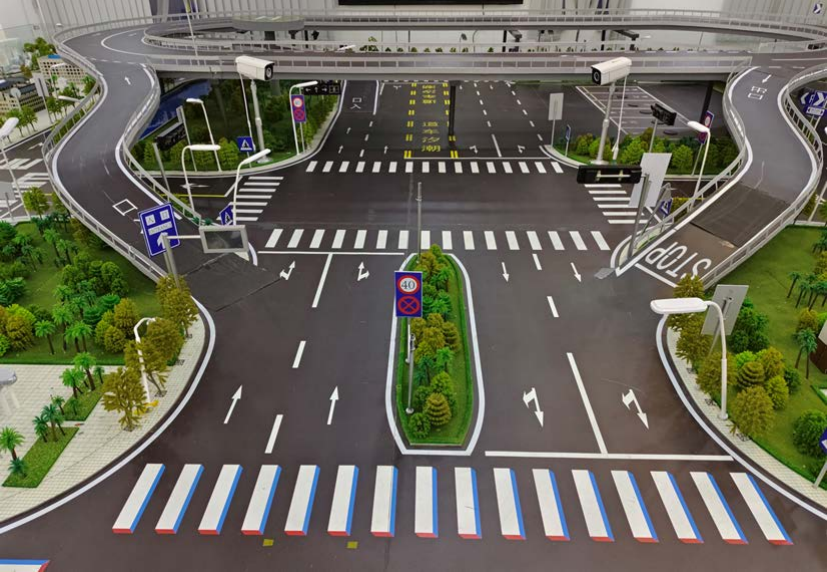}
		\caption{Real MoCAM}
	\end{subfigure}
	\caption{AGX and MoCAM VR platforms}
	\label{platforms}
 \vspace{-0.2in}
\end{figure*}

We implemented the proposed $\mathsf{SVR}$ system using Python in ROS. The virtual world is constructed using CARLA \cite{carla}, which adopts UE for high-performance rendering and PE for fine-grained dynamics modeling. The physical world is connected to CARLA via ROS bridge \cite{carla_bridge} and data sharing is realized via ROS communications, where the nodes publish or subscribe ROS topics that carry the sensory, state, or action information.
The simulation worlds and $\mathsf{IS}^2$ ROS packages are implemented on a Ubuntu workstation with a 
$3.7$\,GHZ AMD Ryzen 5900X CPU and an NVIDIA RTX $3090$\,Ti GPU.
On the other hand, we use car-like robot, LIMO, as vehicles in the physical world, which adopts Ackermann steering.
The robot has onboard IMU, lidar, RGBD camera, and NVIDIA Orin-Nano embedded computing chip for executing the navigation packages. A steering wheel, Logi G29, is used for recording expert driving datasets.
Based on the above hardware and software, we implement two $\mathsf{SVR}$ platforms, i.e., the Agilex (AGX) and Macao Car Racing Metaverse (MoCAM) platforms, as shown in Fig.~\ref{platforms}.

\subsection{Experiment 1: Evaluation of $\mathsf{LIS}$}

\begin{figure*}[!t]
\centering
	\begin{subfigure}{0.245\textwidth}
		\centering
		\includegraphics[height=0.7\linewidth]{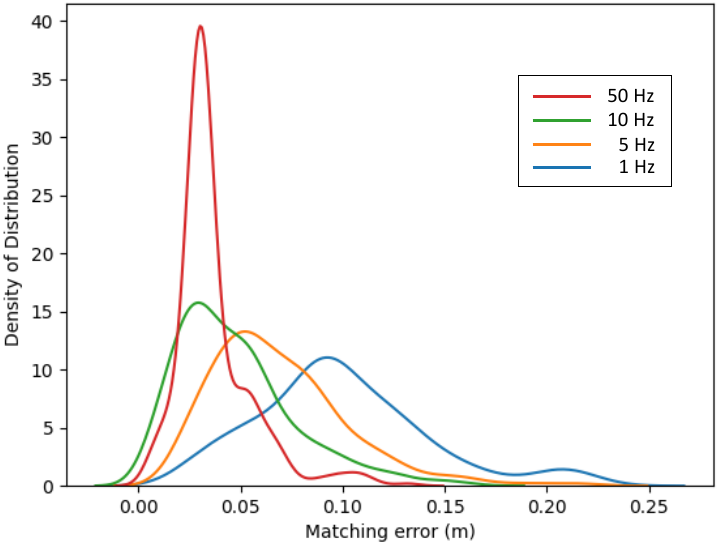}
		\caption{Density of matching errors}
	\end{subfigure}
  	\begin{subfigure}{0.245\textwidth}
		\centering
		\includegraphics[height=0.7\linewidth]{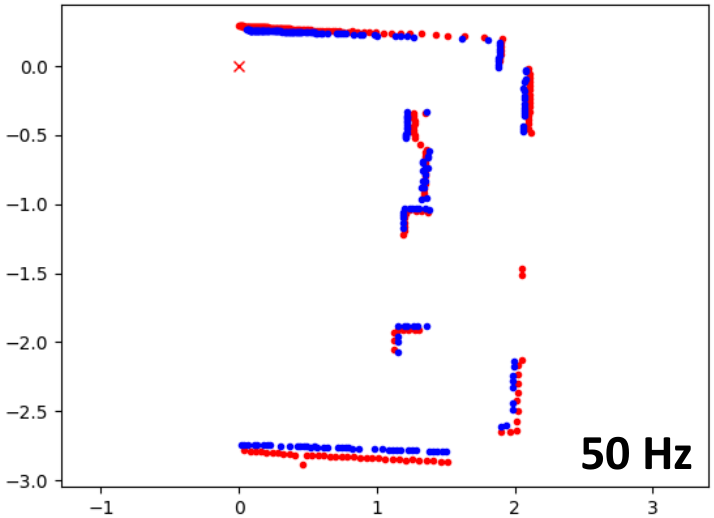}
		\caption{Result at $50\,$Hz}
	\end{subfigure}
  	\begin{subfigure}{0.245\textwidth}
		\centering
		\includegraphics[height=0.7\linewidth]{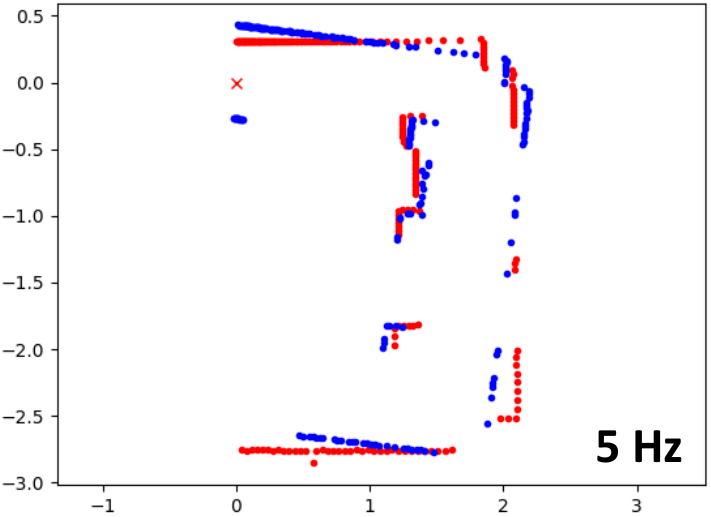}
		\caption{Result at $5\,$Hz}
	\end{subfigure}
 	\begin{subfigure}{0.245\textwidth}
		\centering
		\includegraphics[height=0.7\linewidth]{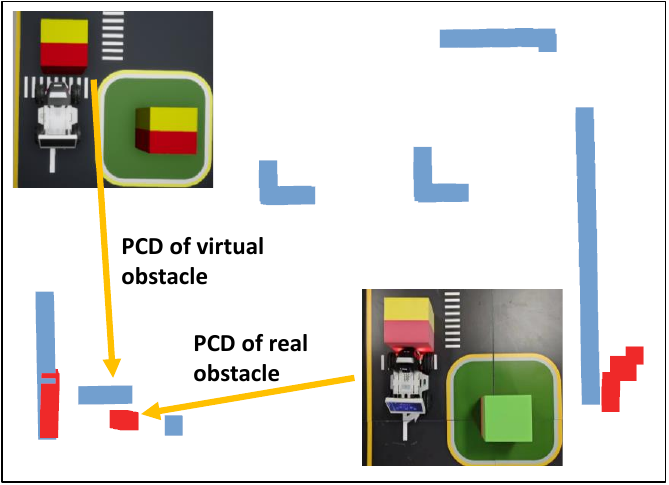}
		\caption{Missed collision due to drift}
	\end{subfigure}
	\caption{Quantitative and qualitative evaluations of the $\mathsf{LIS}$ approach}
	\label{LIS_Evaluation}
 \vspace{-0.2in}
\end{figure*}

We first evaluate the $\mathsf{LIS}$ module in the AGX platform. We use the VR point cloud matching error, which is the average distance between the virtual points and the associated real planes \cite{zhang2014loam}, as a key performance indicator to quantify the accuracy of VR colocation. The probability density of matching error under different colocation frequency is shown in Fig.~\ref{LIS_Evaluation}a. It can be seen that the matching error decreases as the colocation frequency increases. 
Our $\mathsf{LIS}$ method achieves a frequency of $50\,$Hz by using multi-modal measurements, and its average deviation in Fig.~\ref{LIS_Evaluation}a is only $3\,$cm.
This corroborates the qualitative result shown in Fig.~\ref{LIS_Evaluation}b, where the virtual and real point clouds are highly consistent. 
In contrast, when the frequency decreases to $5\,$Hz, the average deviation increases to $>5\,$cm.
This corroborates the qualitative result shown in Fig.~\ref{LIS_Evaluation}c, where the virtual and real point clouds are now inconsistent with each other. 
In fact, the VR inconsistency due to low colocation frequency would become unacceptable in precrash scenarios as shown in Fig.~\ref{LIS_Evaluation}d, resulting in potential missed collisions or model clippings.
This is because the ego vehicle is close to the obstacle in such cases, and a slight delay (e.g., even tens of milliseconds) between two worlds would result in a large difference in virtual and real FoVs and measurements (e.g., red points versus blue points in Fig.~\ref{LIS_Evaluation}d).

\subsection{Experiment 2: Evaluation of $\mathsf{DVSN}$}

We compare our $\mathsf{DVSN}$ to the following benchmarks:
\begin{itemize}
    \item \textbf{No registration}, which directly fuses the VR images.
    \item \textbf{SIFT BF Matcher}, which finds and matches features with the SIFT algorithm \cite{b23} and brute-force matcher.
    \item \textbf{KAZE BF Matcher}, which finds and matches features with the KAZE algorithm \cite{b19} and brute-force matcher.
    \item \textbf{NC-Net}, which utilizes NC-Net to identify feature points. 
\end{itemize}
A basic filter, which solely considers the point distances, is adopted for benchmark approaches to remove outliers. The MAAF is utilized for our $\mathsf{DVSN}$.

The recognizable rate (R-Rate) and object deviation (OD) are adopted to evaluate our $\mathsf{DVSN}$ and the four benchmark schemes.
In particular, the R-Rate is defined as the number of correctly detected objects over the number of all objects \cite{lidarsim}.
This metric is used to quantify the object-level fidelity of synthesized images.
On the other hand, the OD is defined as the relative position shift between the synthesized images and original images, which reflects the geometric fidelity of synthesized images and can be approximated by the mean landmark error \cite{song2017methods}.
In our experiment, we first manually label a set of matched points in each image pair as ground truth. 
Then we calculate the average distance between labeled points and landmarks in virtual and synthesized images, which yields $d_{1}$ and $d_{2}$, respectively. 
Finally, we compute $\mathrm{OD}=|d_1 - d_2| / d_{\mathrm{1}}$. 

The R-Rate and OD results are summarized in Table~\ref{syn_res}. 
The R-Rate of our approach is $98.8$\%, which is close to $100$\% and higher than all the other schemes. 
The average OD of our method is merely $3.2$\%. 
Furthermore, the percentage of high quality synthesized images (i.e., with OD $<5$\%) under our approach is $85.7$\%, which is at least $8$\% higher than all the other schemes. 
The percentage of low quality synthesized images (i.e., with OD $>10$\%) is $5.7$\%, which is also the smallest among all the evaluated schemes. 
Lastly, it is found that NC-Net outperforms conventional non-deep-learning registration methods, and the proposed MAAF can further improve the R-Rate performance. 
Note that the R-Rate of no registration scheme and the OD of the SIFT BF matcher scheme are omitted, as these metrics are not applicable to the two schemes.

\begin{table}[!t]
\caption{Evaluation results of $\mathsf{DVSN}$}
\vspace{-0.05in}
\begin{center}
\scalebox{0.9}{
\begin{tabular}{ccccc}
\hline
 & \textbf{R-Rate} &{\textbf{Avg. OD}} & \textbf{OD \textless 5\%} & \textbf{OD \textgreater 10\%} \\
\hline
SIFT + BF matcher & 47.5\% & - & - & -\\
No Registration & - & 10.9\% & 14.3\% & 45.7\% \\
KAZE + BF Matcher & 84.7\% &4.9\% &42.9\% & 5.7\% \\
NC-Net & 97.7\% &3.8\% &77.3\% & 8.6\%  \\
\hline
\textbf{NC-Net with MAAF (ours)} &  \textbf{98.8}\textbf{\%} & \textbf{3.2}\textbf{\%}& \textbf{85.7\%} & 5.7\%\\
\hline
\end{tabular}
}
\label{syn_res}
\end{center}
\vspace{-0.1in}
\end{table}

\subsection{Experiment 3: Evaluation of $\mathsf{SVR}$-Trained IL}

\begin{figure}
\centering
\includegraphics[width=0.98\linewidth]{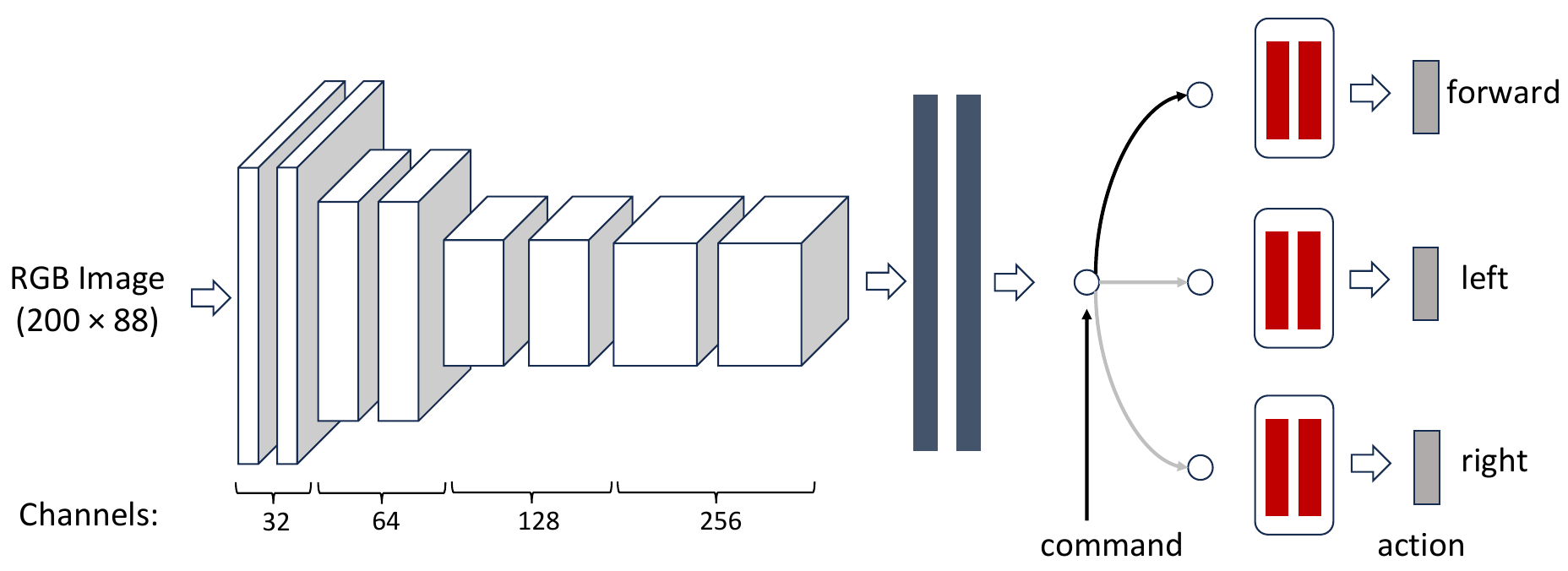}
\caption{Structure of the end-to-end IL network.}
\label{CIL}
\end{figure}

\begin{figure*}[!t]
	\centering 
 	\begin{subfigure}{0.325\linewidth}
 \vspace{1mm}
		\centering
		\includegraphics[width=1\linewidth,height=0.95\linewidth]{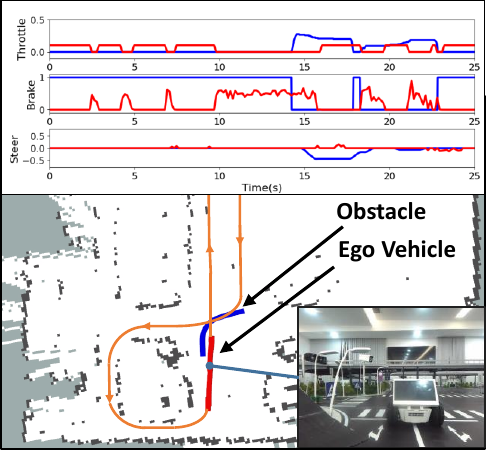}
    \caption{Trajectories and actions of ego and obstacle vehicle in car-following scenario.}
	\end{subfigure}
	\begin{subfigure}{0.325\linewidth}
		\centering
		\includegraphics[width=1\linewidth,height=0.95\linewidth]{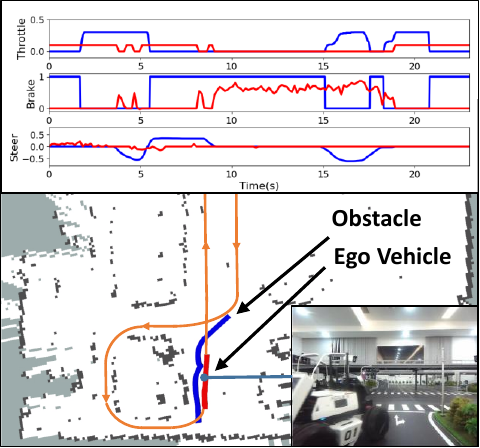}
  \caption{Trajectories and actions of ego and obstacle vehicle in cut-in scenario.}
	\end{subfigure}
 \centering
	\begin{subfigure}{0.325\linewidth}
 \vspace{1mm}
		\centering
		\includegraphics[width=1\linewidth,height=0.95\linewidth]{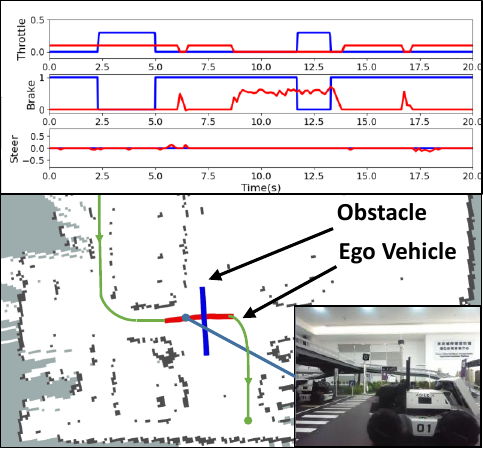}
    \caption{Trajectories and actions of ego and obstacle vehicle in running-red-light scenario.}
	\end{subfigure}
	\caption{Real-world evaluation results of $\mathsf{SVR}$-trained $\mathsf{IL}$ in car-following, cut-in, running-red-light scenarios.}
	\label{closed-loop}
 \vspace{-0.05in}
\end{figure*}

To verify the performance gain brought by $\mathsf{SVR}$ to AD, this experiment trains the driving model on the dataset collected in VR space and tests the 
$\mathsf{SVR}$-trained model in real-world environments.
We collect $14000$ (image, action) data samples, with $7000$ real samples and $7000$ virtual samples, and generate another $7000$ synthesized samples.
The entire dataset is then categorized into the following groups:
\begin{itemize}
        \item \textbf{Real dataset}, which only contains images and actions from the real environment (without obstacle vehicles).
        \item \textbf{Virtual dataset}, which only contains images and actions from the virtual environment (with obstacle vehicles).
        \item \textbf{No-Registration synthesized dataset}, which contains directly synthesized images and real-world actions.
        \item \textbf{$\mathsf{SVR}$ dataset}, which involves image registration and contains synthesized images and real-world actions.
        \item \textbf{Enhanced $\mathsf{SVR}$ dataset}, which contains all the $\mathsf{SVR}$ dataset and background-changed data (\textless 5\%) generated by the simulator.
\end{itemize}
A human-in-the-loop interface is developed to allow driving expert to both see and interact with virtual content that is superimposed over the real world, which can unlock more natural human–robot interaction as shown in our video. 
The above datasets are separately fed to the IL network\footnote{
The high-level command $c\in \{\mathrm{forward}, \mathrm{left}, \mathrm{right}\}$. The output action $\mathbf{a}$ is a three-dimensional vector that controls steer $s$, brake $b$, and throttle $t$: $\mathbf{a}=\langle s,b,t \rangle$. Given human expert's action $\mathbf{a}_{\mathrm{gt}}= \langle s_{\mathrm{gt}}, b_{\mathrm{gt}},t_{\mathrm{gt}}\rangle $, the training loss function is $\mathcal{L} (\mathbf{a}, \mathbf{a}_{\mathrm{gt}})=\lambda_s\left \| s - s_{\mathrm{gt}} \right \|^{2}_2 + \lambda_b\left \| b - b_{\mathrm{gt}} \right \|^{2}_2 + \lambda_t\left \| t - t_{\mathrm{gt}} \right \|^{2}_2$.}
shown in Fig.~\ref{CIL} for model training. The IL network is implemented according to \cite{cil}, 

\begin{table*}[!t]
\centering
\caption{Comparison of $\mathsf{IL}$ models trained by different datasets}
\vspace{-0.15in}
\begin{center}
\begin{tabular}{ccccccc}
\hline
\multirow{2}{*}{} & \multirow{2}{*}{\#\,intervention} & \multirow{2}{*}{\#\,missed-turn} & \multicolumn{4}{c}{collision avoidance success rate}\\
\cline{4-7}
&&&  cut-in & car-following & running the red light & average \\
\hline
virtual dataset & 3.33& 66.7\%& 0.1& 0.4& 0.2& 0.23\\
no-registration synthesized dataset & 1& 11.1\%& 0.5& 0.6& 0.5& 0.53\\
$\mathsf{SVR}$ dataset (ours)& 0.67& 11.1\%& 0.6& 0.7& 0.7& 0.67\\
\hline
enhanced $\mathsf{SVR}$ dataset (ours) & \textbf{0.33 }& \textbf{5.6\% }& \textbf{0.9}& \textbf{1.0}& 0.7& \textbf{0.87}\\
\hline
\end{tabular}
\label{tab4}
\end{center}
\end{table*}

\textbf{Metrics and Scenarios}: The number of interventions and missed-turns are used to evaluate the performance of IL models trained by different datasets \cite{cil}. 
In particular, we intervene the vehicle whenever it moves off the road (i.e., regarded as one intervention), and manually steer the vehicle whenever it fails to make the turn at intersections (i.e., regarded as one missed-turn). 
We test the trained IL models on the LIMO vehicle in the MoCAM platform and the two target routes (marked in orange and green) are shown in Fig.~\ref{closed-loop}. Each route is tested 6 times. 
The dynamic obstacle vehicle (also LIMO), is controlled by a human expert and challenge the IL-driven vehicle according to 3 critical scenarios defined by NHTSA (https://www.nhtsa.gov/), i.e., car following, cut in, running red lights, with each scenario being tested 10 times.
These challenges, together with the obstacle vehicle, are not within the real training datasets. The IL-driven vehicle is expected to accomplish the routes autonomously while avoiding collisions with any other obstacles (including both the static roads and the dynamic obstacle vehicle). Consequently, the success rate of collision avoidance is also a key metric for evaluation.

\begin{figure}[!t]
\centering
\includegraphics[width=1\linewidth]{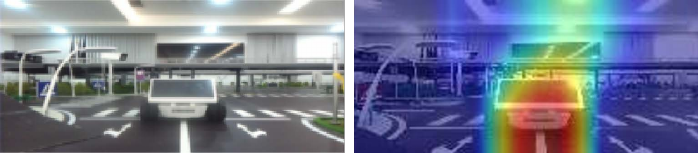}
\caption{Activation map of IL in the car-following scenario}
\label{feature_map}
\end{figure}

\textbf{Results}: The trajectories and actions of the ego and obstacle vehicles under the proposed $\mathsf{SVR}$-trained IL are shown in Fig.~\ref{closed-loop}, where the red trajectories are generated by the IL model and the blue trajectories are generated by the human challenger. The associated throttle, steer, and brake commands are shown in the higher part of Fig.~\ref{closed-loop}. 
First, for the car-following scenario in Fig.~\ref{closed-loop}a, 
the obstacle vehicle is stopping in front of the traffic light from $t=0\,$s to $t=14\,$s. 
The IL-driven vehicle approaches the obstacle vehicle using a go-stop policy that generates braking signals intermittently, so as to make the best prompt reactions whenever the obstacle vehicle moves. 
This corroborates the activation map of IL shown in Fig.~\ref{feature_map}, which is obtained by the Grad-CAM method \cite{8237336}.
It can be seen that the attention layers marked in red are highly focused on the obstacle vehicle, even if the IL model has never ``seen'' the real LIMO before in the physical world. 
This is because the IL model learns from the VR data generated by our proposed $\mathsf{SVR}$ platform. As such, the navigation efficiency and safety are both guaranteed.
Next, we consider the cut-in scenario shown in Fig.~\ref{closed-loop}b.
Starting from $t=0\,$s, the IL-driven vehicle navigates autonomously at the right lane along the orange target path.
At $t=1.5\,$s, the obstacle vehicle at the left lane accelerates to the maximum speed, turns right at $t=3.5\,$s, and after cutting in to the right lane it suddenly brakes at $t=5.5\,$s. 
The IL-driven vehicle reacts to the challenger properly and continues its navigation task after the challenger leaves. 
Finally, we consider the running-red-light scenario shown in Fig.~\ref{closed-loop}c. 
In this experiment, the obstacle vehicle breaks the red light at the intersection when the ego vehicle is crossing the same intersection legally. Again, the IL-driven vehicle can handle this unseen situation by leveraging its knowledge learnt from the VR space. 

\begin{figure*}[!t]
\centering
\includegraphics[width=1\linewidth]{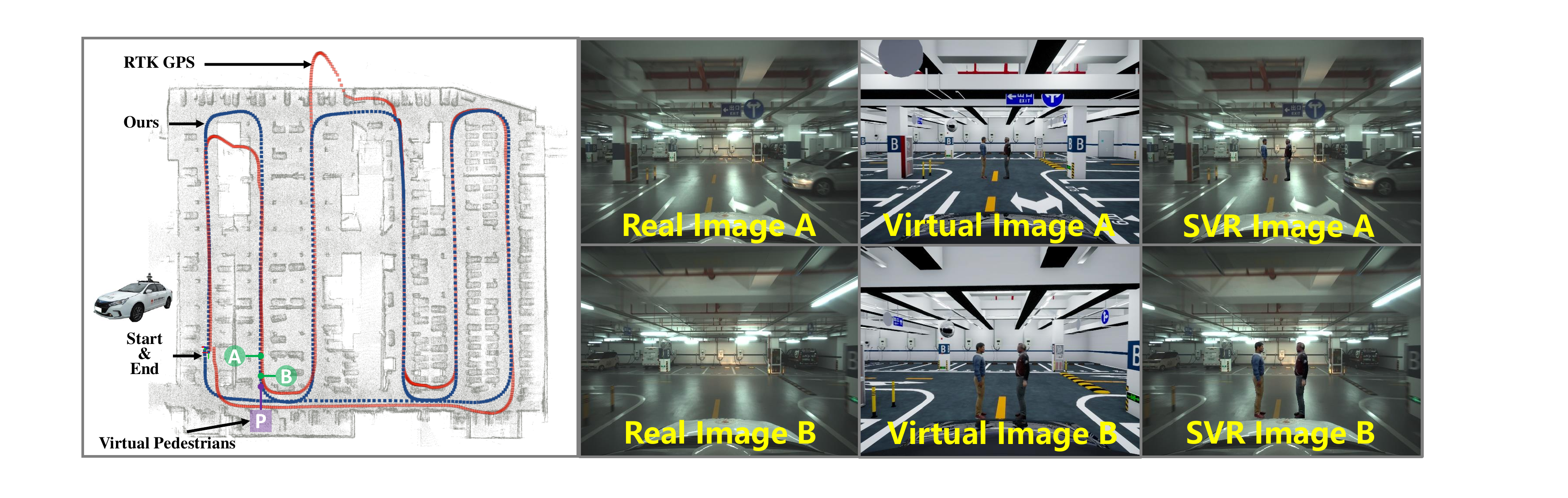}
\caption{Trajectory (marked in blue) and images (on the right hand side) obtained from $\mathsf{SVR}$. Virtual, real, and $\mathsf{SVR}$ images are collected at green circles A and B, respectively. Virtual pedestrians stand at the purple square, talking with each other. For comparison, trajectory obtained from the RTK-GPS is marked in red.}
 \label{real_world}
\end{figure*}

The quantitative results are presented in Table~\ref{tab4}. 
First, IL trained by virtual dataset leads to the worst performance in all metrics due to domain shifts. Compared to other schemes, its number of interventions or missed-turns is at least twice larger. In contrast, IL trained by real dataset can reduce the number of interventions or missed-turns. However, it always collides with the obstacle vehicle in the test track in our experiments (hence is omitted in Table~\ref{tab4}). These observations demonstrate the necessity of employing VR in AD.
Second, by comparing the no-registration and the $\mathsf{SVR}$ schemes, it can be seen that ignoring registration degrades the collision avoidance capability of IL, as there exists deviation during synthesis.
Third, the $\mathsf{SVR}$ dataset increases the success rate for collision avoidance by at least $10\%$ compared to previously mentioned schemes in the aforementioned adversarial scenarios. 
This demonstrates the necessity of employing $\mathsf{IS}^2$ design in VR systems. 
Lastly, the enhanced $\mathsf{SVR}$ dataset, which introduces background variation into our method would further improve the performance. 
This is because background variation would make the IL model more focused on foreground dynamic obstacles and less sensitive to background environments, which corroborates the observations in \cite{DR}. 

\subsection{Experiment 4: Evaluation of $\mathsf{SVR}$ on Passenger Vehicle}

Finally, we validate the robustness of $\mathsf{SVR}$ on a passenger autonomous vehicle in a parking lot scenario.
The vehicle is equipped with $6$ cameras, a $128$-line 3D lidar, an RTK GPS, and an onboard Intel x86 computer. 
It can be seen from Fig.~\ref{real_world} that the trajectory (marked in blue) obtained from $\mathsf{SVR}$ matches the parking lot topology very well. 
This makes it possible for $\mathsf{SVR}$ to seamlessly merge virtual pedestrians into real parking lot accurately and smoothly, resulting in a naturalistic pre-crash VR testing scenario as shown in real, virtual, and SVR images at positions A and B of Fig.~\ref{real_world}.
In contrast, there exist significant deviations for the RTK-GPS trajectory (marked in red) \cite{dense-rl}.
For instance, at the upper side of the map in Fig.~\ref{real_world}, the RTK-GPS trajectory is outside the parking lot, making the virtual vehicle crash into walls.

\section{Conclusion}
This paper presented a high-fidelity $\mathsf{SVR}$ platform for AD. This platform was driven by the $\mathsf{IS}^2$ technique, which consists of $\mathsf{LIS}$ to map autonomous vehicles in the real world and adversarial obstacles in the virtual world to a symbiotic world, and $\mathsf{DVSN}$ to calibrate and aggregate the synchronized VR sensor and action data. Various experiments were conducted, which prove the necessity of $\mathsf{SVR}$ for AD and the superiority of $\mathsf{IS}^2$ over other benchmarks. 
By integration with large multi-modal models (e.g., gemini, sora), the developed $\mathsf{SVR}$ platform opens up opportunities for metaverse autonomous driving.

\ifCLASSOPTIONcaptionsoff
  \newpage
\fi

\bibliographystyle{IEEEtran}
\bibliography{li_svr_ral}
\end{document}